\documentclass[10pt,twocolumn,letterpaper]{article}

\usepackage[pagenumbers]{cvpr} 

\definecolor{cvprblue}{rgb}{0.21,0.49,0.74}
\usepackage[pagebackref,breaklinks,colorlinks,allcolors=cvprblue]{hyperref}

\usepackage{multirow}
\usepackage{amsmath}
\usepackage{array}
\usepackage{booktabs} 
\usepackage{rotating} 
\newcommand*{\affaddr}[1]{#1} 
\newcommand*{\affmark}[1][*]{\textsuperscript{#1}}
\newcommand\ZLChange[1]{{#1}}

\usepackage[accsupp]{axessibility}

\def\paperID{31226} 
\def\confName{CVPR}
\def\confYear{2026}

\makeatletter 
\def\thanks#1{\protected@xdef\@thanks{\@thanks\protect\footnotetext{#1}}}
\makeatother
\title{Dance Across Shifts: Forward-Facilitation Continual Test-Time Adaptation through Dynamic Style Bridging}

\author{
    Zhilin Zhu\affmark[1,2], 
    Yabin Wang\affmark[1], 
    Zhiheng Ma\affmark[3,4], 
    Yaguang Song\affmark[2],
    Yaowei Wang\affmark[1,2],
    Xiaopeng Hong\affmark[1,2,$\ast$] \\
    \affaddr{\affmark[1]Harbin Institute of Technology}\quad
    \affaddr{\affmark[2]Pengcheng Laboratory}\\
    \affaddr{\affmark[3]Shenzhen University of Advanced Technology}\\
    \affaddr{\affmark[4]Guangdong Provincial Key Laboratory of Computility Microelectronics}\\
    {\tt\small \{zhuzhl, songyg01, wangyw\}@pcl.ac.cn, }\\
    {\tt\small wangyabin@hit.edu.cn, mazhiheng@suat-sz.edu.cn, hongxiaopeng@ieee.org}
}
\thanks{{$\ast$ Corresponding author}}

\begin{document}


\maketitle
\begin{abstract}
Continual Test-Time Adaptation (CTTA) aims to empower perception systems to handle dynamic distribution shifts encountered after deployment. Existing methods predominantly follow a backward-alignment paradigm, which rigidly aligns incoming data with supervisory surrogates derived from the source domain. Consequently, they struggle with unreliable supervision and evolving distribution shifts. To overcome these limitations, we introduce a novel forward-facilitation paradigm through a method termed Dynamic Style Bridging. Prior to deployment, we construct a compact knowledge base of generated class exemplars. During test time, to mitigate inherent generative bias and adapt these proxies to incoming data, we propose a multi-level bridging mechanism. This mechanism dynamically injects the proxies with incoming data styles at the input, statistical, and representation levels, while preserving the original semantics of the proxies. These high-fidelity proxies are then used to provide reliable, on-demand supervisory signals, enabling stable adaptation under continual shifts. Extensive experiments across standard CTTA benchmarks demonstrate that our method achieves consistent and substantial improvements over recent state-of-the-art approaches. Code is available at \href{https://github.com/z1358/DAS}{https://github.com/z1358/DAS}.
\end{abstract}    
\section{Introduction}
\label{sec:intro}
Deep learning models have achieved remarkable success in controlled environments where training and test data distributions {remain stationary~\cite{dosovitskiyimage,huang2024rtracker,xie2024d3still}.} However, real-world scenarios are inherently dynamic, characterized by ongoing distribution shifts and non-stationary data streams~\cite{niu2021adaxpert,xiao2024beyond,lu2024visual}. Such distributional variations arise across diverse application domains and substantially undermine the reliability of static pre-trained models~\cite{koh2021wilds,sakaridis2021acdc,tan2024uncertainty,deng2025multi}. Addressing this challenge requires learning systems capable of continual adaptation to evolving conditions after deployment.

\begin{figure}[t]
\centering
\includegraphics[width=0.96\columnwidth]{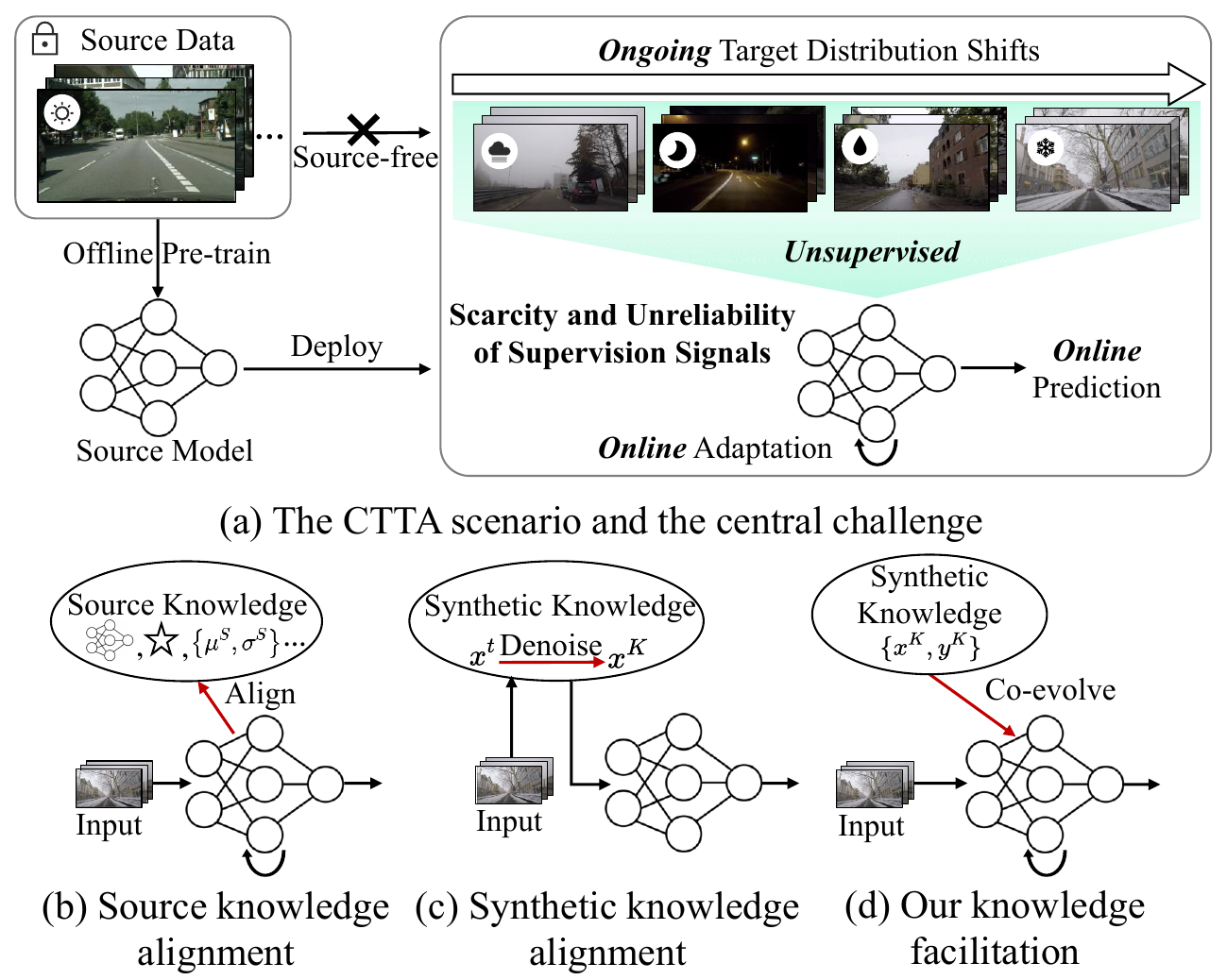} 
\caption{Illustration of the considered CTTA problem and comparison of different frameworks. (a) The pipeline and the central challenge of CTTA. (b)-(c) Existing methods primarily focus on the \emph{backward-alignment} paradigm. (d) Our approach explores a completely different \emph{forward-facilitation} paradigm. By co-evolving with the distribution, we continually transform static synthetic knowledge to the current target domain, directly addressing the central challenge.}
\label{settings_and_piplines}
\end{figure}

This demand has spurred the emergence of continual test-time adaptation (CTTA)~\cite{Wang_2022_CVPR,gan2023decorate,zhao2025d,yang2024versatile,press2024rdumb}, which provides a promising direction for mitigating the domain shift problem after model deployment. Here, an off-the-shelf model is adapted to continually changing target domains in an online manner, without access to its original training data due to privacy constraints and without altering its initial training procedure. As illustrated in ~\cref{settings_and_piplines} (a), the unsupervised and online nature of the test phase introduces a fundamental dilemma: \emph{the scarcity and unreliability of supervision signals}. This challenge is further exacerbated in scenarios involving continually shifting distributions, leading to intensified error accumulation and catastrophic forgetting, both of which severely hinder stable and effective model adaptation~\cite{gong2023sotta,liang2025comprehensive,vuong2025preserving,chen2025reducing,Wang_2024_WACV}.

To address these limitations, existing methods have predominantly followed a \emph{backward-alignment} paradigm. As illustrated in~\cref{settings_and_piplines} (b),  whether through parameter regularization~\cite{niu2022efficient,Brahma_2023_CVPR,ijcai2023p183}, representation alignment~\cite{Dobler_2023_CVPR,chakrabarty2023santa,li2025continual}, or statistical prior matching~\cite{wang2024distribution,yoo2024and,Zhang2025DPCoreDP}, these methods attempt to construct supervisory surrogates from source-domain knowledge to guide test-time adaptation. However, such surrogates are inherently weak approximations of unavailable ground-truth supervision, and this backward-alignment paradigm rigidly forces a comparison between the model's current, evolving state (computed on the unsupervised target data) and these weak, static anchors. As a result, the model is often driven by noisy or outdated alignment cues, undermining stable and effective adaptation under continual shifts. Recent advances in generative diffusion models~\cite{nie2022diffusion,tsai2024gda} have inspired methods such as DDA~\cite{gao2023back} and SDA~\cite{guo2025everything}, which project all inputs back to a synthetic domain at test time. As depicted in ~\cref{settings_and_piplines} (c), such approaches still force alignment of changing target data to a static anchor (synthetic domain), fail to provide reliable supervision, and are laden with generative bias~\cite{guo2025everything}. Whether relying on weak source surrogates or costly generative denoising, the backward-alignment paradigm struggles to elegantly and directly address the core challenges of CTTA.

Departing from the prevailing pursuit of supervisory surrogates, we confront this core challenge from first principles~\cite{irwin1990aristotle} and propose a novel \emph{forward-facilitation} paradigm. Our central thesis is that instead of backwardly constraining the model with tenuous, static knowledge, we should actively evolve reliable knowledge to the present context to provide tailored supervision. To this end, we introduce a dynamic style bridging framework designed to directly overcome the aforementioned limitations through two complementary components. First, leveraging advanced generative models, we construct a compact offline knowledge base of semantically pure class exemplars. This serves as an explicit and trustworthy semantic foundation that the backward-alignment paradigm fundamentally lacks. Second, to avoid generative bias and bridge the dynamic domain gap, our multi-level bridging mechanism abandons rigid alignment with static anchors and instead actively adapts this reliable knowledge to match the style of the current target batch. By injecting target-specific style at the input, statistical, and representation levels, the knowledge base becomes a time-varying one that co-evolves with data streams. As illustrated in ~\cref{settings_and_piplines} (d), this process generates semantically reliable and aligned supervision on-the-fly, promoting discriminative adaptation with on-demand guidance. In summary, our contributions are as follows:

\begin{itemize}
    \item We introduce a novel {forward-facilitation} paradigm for CTTA, marking a fundamental shift from the prevailing backward-alignment strategies. Our approach redefines the role of prior knowledge from a static constraint into a dynamic asset that actively facilitates adaptation.
    \item We propose a dynamic style bridging framework to realize this paradigm. By integrating a semantically explicit synthetic knowledge base with a multi-level bridging mechanism that customizes it to the current target style, our method directly resolves the core CTTA dilemma of scarce and unreliable supervision.
    \item Extensive experiments across challenging benchmarks demonstrate that our framework consistently and substantially outperforms existing state-of-the-art methods, highlighting its effectiveness and generalizability.
\end{itemize}

\section{Related Work}
\begin{figure*}[!th]
\centering
\includegraphics[width=0.92\textwidth]{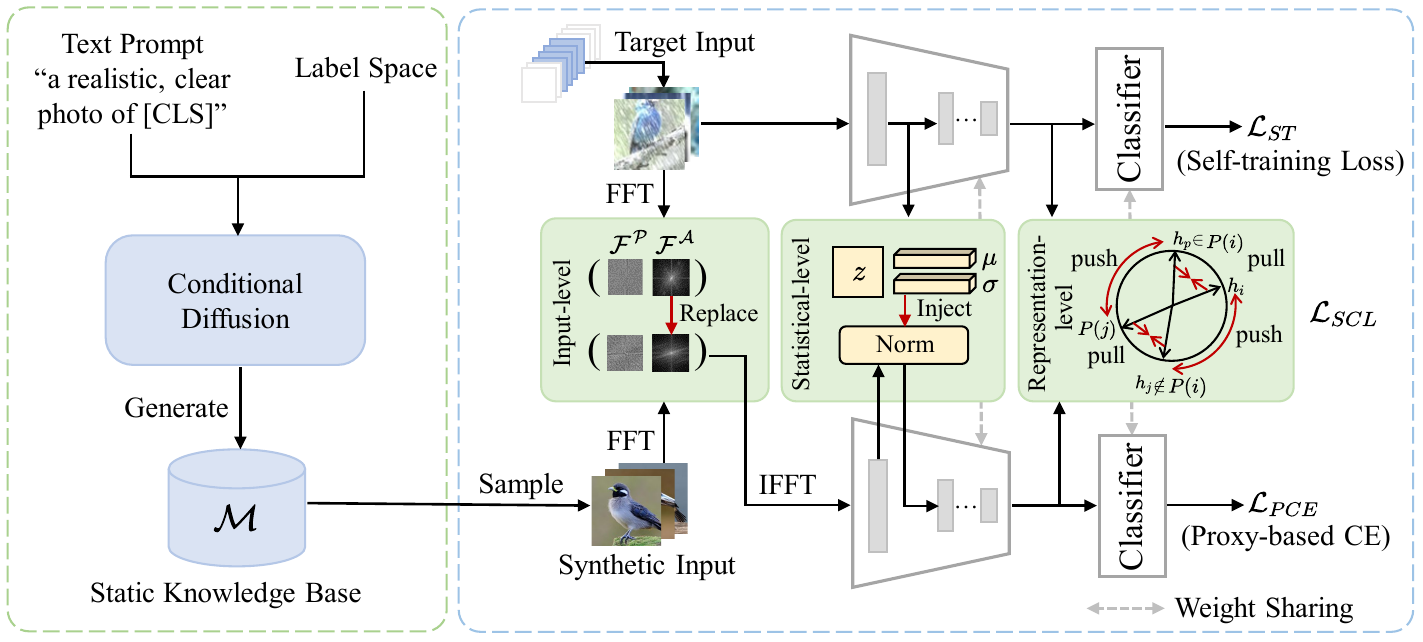} 
\caption{The illustration of our framework. We construct in advance a compact set of proxies containing synthetic knowledge that encapsulates explicit semantic information. During the CTTA process, our proposed multi-level bridging mechanism dynamically transforms the static knowledge in accordance with the evolving data stream, precisely delivering the reliable supervision signals required for the model adaptation, thereby robustly supporting the forward-facilitation paradigm.}
\label{fig:framework}
\end{figure*}
\subsection{Test-Time Adaptation}
{TTA addresses distribution shifts by adapting pre-trained models to an unlabeled target domain during inference without access to source data~\cite{chen2022contrastive,niu2024test,huang2025cosmic,ma2025surgeon}.}
Early methods~\cite{sun2020test,liu2021ttt} rely on self-supervised proxy tasks to derive adaptation signals. To avoid altering the training process or model architecture, subsequent methods have explored objectives such as entropy minimization~\cite{wang2021tent,niu2023towards,zhou2025test} and consistency regularization~\cite{zhang2022memo,nguyen2023tipi,wang2023feature}. While TTA primarily addresses adaptation to a single static target domain, practical scenarios often involve test data streamed from non-stationary and evolving environments~\cite{yao2024socialized,Wang_2022_CVPR,zhou2024class,chen2026spectral}. Consequently, our focus is on online continual adaptation, which is more consistent with real-world demands.

\subsection{Continual Test-Time Adaptation}
CTTA is an emerging research area that has received increasing attention due to its practicality. CoTTA~\cite{Wang_2022_CVPR} employs a teacher-student framework to generate pseudo-labels for model updating and randomly recovering model parameters back to the source model to preserve source-domain knowledge. 
A series of subsequent works has explored various source knowledge alignment strategies, encompassing data-driven parameter regularization~\cite{Brahma_2023_CVPR,niloy2024effective}, representative feature alignment~\cite{Dobler_2023_CVPR,chakrabarty2023santa,yoo2024and}, source-domain relationship bootstrapping~\cite{kang2023leveraging,zhu2024reshaping}, and statistical information matching~\cite{chen2024each,Zhang2025DPCoreDP,wang2025efficient}. Meanwhile, some methods mitigate catastrophic forgetting by freezing the model and introducing additional parameters, but they typically require extensive pre-adaptation warm-up using a large amount of source data~\cite{song2023ecotta,liu2024vida,lee2024becotta,li2025continual}. Beyond these source-domain knowledge explorations, recent approaches have attempted to utilize synthetic knowledge to adapt samples with distribution shifts at the input level~\cite{gao2023back,guo2025everything}. However, these approaches inevitably incur computationally expensive overhead at test time, greatly diminishing their practicality in real-world applications. {Instead, we tailor synthetic knowledge to evolving target domains through a novel forward-facilitation paradigm. This design ensures both adaptation effectiveness and computational efficiency, making it well-suited for real-time scenarios.}

\subsection{Synthetic Knowledge for Discriminative Tasks}
Beyond information gathered from real-world data, synthetic content generated by generative models has demonstrated substantial potential in enhancing visual representations across a wide range of discriminative tasks~\cite{tian2023stablerep,fan2024scaling,wu2025synthetic}. Recently, there has been a growing interest in exploiting the extensive synthetic knowledge embedded within generative models. 
In contrast to prior work that uses synthetic data for training-time augmentation~\cite{azizi2023synthetic,hammoud2024synthclip, Tian_2024_CVPR, zhang2025revisiting}, our method emphasizes the continual evolution of synthetic knowledge to generate supervisory signals during the deployment phase.

\section{Methodology}
\subsection{Preliminaries and Overall Framework}
Given a model $f_\theta$ pre-trained on the labeled source domain $\mathcal{D}^{S}$, CTTA~\cite{Wang_2022_CVPR,gan2023decorate,ni2025maintaining,wang2025paid} aims to adapt it to a sequence of unlabeled and dynamically evolving target domains $\left [ \mathcal{D}^{T_1}, \mathcal{D}^{T_2},\cdots, \mathcal{D}^{T_n}\right ]$ during test time. Each target domain shares the same label space as the source domain. Note that the model is unaware of when domain shifts occur, and in practical deployment, the number of target domains may be unknown or even unbounded. The model operates in an online manner, receiving a stream of test batches $\left \{ B_t \right \} _{t=1}^\infty $, and performing adaptation incrementally by processing one target batch $B_t$ at each time step $t$.

Our framework is illustrated in \cref{fig:framework}. 
Leveraging recent advancements in text-to-image synthesis, our method integrates a compact base of synthetic knowledge enriched with explicit semantic content into the CTTA pipeline. To effectively handle evolving test-time distributions, we propose a multi-level bridging mechanism that customizes the synthetic knowledge to the current target domain at multiple levels, thereby delivering precise and informative supervision signals for ongoing adaptation. Following previous works~\cite{Dobler_2023_CVPR,zhu2024reshaping,ni2025maintaining}, a self-training loss is applied to ensure stability during adaptation. The components are detailed in the following subsections.

\subsection{Synthetic Knowledge Establishment}
As shown in ~\cref{settings_and_piplines} (a), the central challenge in CTTA lies in the absence of reliable supervision signals, which often forces models to rely on their own predictions (i.e., pseudo-labels) for self-training. However, such supervision is inherently noisy, especially under domain shifts, leading to error accumulation and catastrophic forgetting~\cite{zhang2025dca,wang2023isolation,fan2024dynamic,zhu2025revisiting,wang2025effortless}. To break this cycle, we construct a semantically pure and stable knowledge base that serves as a dependable source of supervision throughout the adaptation process.

Inspired by the remarkable capabilities of text-to-image generative models, we construct this knowledge base offline through a one-time pre-processing step. For a dataset comprising $C$ classes, we utilize a pre-trained diffusion model (e.g., Stable Diffusion~\cite{rombach2022high}) to synthesize a compact set of class-representative images, characterized predominantly by a single object per image. Using descriptive text prompts such as ``a realistic, clear photo of $[CLS]$, on a clean background", we generate a small set of $M$ prototypical exemplars for each class. This process yields a compact synthetic knowledge base, denoted as $\mathcal{M}={(x_i^K,y_i^K)}_{i=1}^{C\times M}$, where $x_i^K$ is a synthetic image and $y_i^K$ is its corresponding class label. The resulting $\mathcal{M}$ possesses two crucial advantages:

\noindent \textbf{Semantic purity}: Each synthetic image serves as a prototypical class exemplar, free from real-world confounders like background clutter or occlusions, making it an ideal semantic anchor. 

\noindent \textbf{Computational efficiency}: Pre-generation eliminates the substantial overhead of invoking large diffusion models during online adaptation, ensuring practical deployment feasibility. This lightweight design establishes a solid foundation for subsequent dynamic style bridging operations.

\subsection{Multi-Level Bridging Mechanism}
A central characteristic of real-world deployment is the evolving nature of test-time environments. The presence of changing distribution shifts renders conventional backward-alignment approaches suboptimal, as they often establish a tenuous trade-off that stifles effective adaptation~\cite{zhou2024test,zhu2024reshaping}. In contrast, we depart from the static anchor philosophy, proposing a novel forward-facilitation paradigm instead. Rather than constraining the model with static past knowledge, we adaptively transform our synthetic knowledge base $\mathcal{M}$ to align with the style of the incoming target batch, projecting existing knowledge into the present context.

To achieve this, we propose a multi-level bridging mechanism that progressively adapts synthetic knowledge to the current target domain, operating hierarchically from appearance to representation. All operations are performed jointly on a randomly sampled synthetic batch $B_K$ and the current target batch $B_t$. Samples from the two batches are paired in index order purely for implementation convenience, ensuring batch-level style alignment. For clarity, we illustrate the process using a single image pair $(x_i^K, x_j^t)$, while all operations and losses are computed over mini-batches in practice. At the input level, we first perform visual style injection~\cite{yang2020fda} via Fast Fourier Transform (FFT). Exploiting the Fourier Transform's capacity to decouple image style (amplitude spectrum $\mathcal{F}^\mathcal{A}$) from content (phase spectrum $\mathcal{F}^\mathcal{P}$), we efficiently inject the target style through amplitude spectrum replacement. Let $\mathcal{F}(\cdot)$ and $\mathcal{F}^{-1}(\cdot)$ denote the FFT and its inverse, respectively. For each synthetic-target image pair $(x_i^K, x_j^t)$, the stylized image $\tilde{x}_i^K$ is formulated as:
\begin{equation}
    \tilde{x}_i^K = \mathcal{F}^{-1}([\mathcal{F}^\mathcal{A}(x_j^t), \mathcal{F}^\mathcal{P}(x_i^K) ]).
    \label{eq:amplitude_spectrum_replacement}
\end{equation}
This operation ensures that synthetic exemplars acquire target domain stylistic characteristics at the input level prior to encoder processing. Subsequently, at the shallow feature level, we bridge style statistics inspired by neural style transfer methods~\cite{huang2017arbitrary}. We extract instance-level statistics (mean $\mu$ and standard deviation $\sigma$) from a low-layer feature map $z$ for both the stylized synthetic sample and the target sample. These statistics are computed channel-wise across spatial dimensions. The feature representation of the knowledge sample, $z(\tilde{x}_i^K)$, is then transformed by aligning its statistics with those of the target sample:
\begin{equation}
\tilde{z}(\tilde{x}_i^K) = \sigma_j^t \left( \frac{z(\tilde{x}_i^K) - \tilde{\mu}_i^K}{\tilde{\sigma}_i^K} \right) + \mu_j^t.
\label{eq:adain_bridge}
\end{equation}
This operation normalizes the style of the knowledge features to match the first- and second-order statistics of the target features, enforcing consistency at the shallow feature level. Finally, we perform contrastive learning~\cite{zhu2022multi} at the semantic representation level to minimize discrepancies between stylized synthetic data and target samples. Pseudo-labels are generated for target samples based on model predictions, and the supervised contrastive loss is applied as:
\begin{equation}
\mathcal{L}_{SCL} = -\sum_{i} \sum_{p \in P(i)} \log \frac{\exp(\text{sim}(h_i, h_p))}{\sum_{j} \exp(\text{sim}(h_i, h_j))},
\label{eq:scl_bridge}
\end{equation}
where $h_i$ denotes the representation of the $i$-th sample, $P(i)$ refers to the set of positive samples within the joint batch sharing the same label, and $\text{sim}(\cdot,\cdot)$ is the cosine similarity. This objective clusters same-class samples across domains in the representation space, thereby completing deep semantic-level alignment.

\begin{table*}[th]
\caption{Classification error rate (\%, lower is better) for the standard CTTA task on ImageNet-to-ImageNetC. All results are evaluated at the highest corruption severity level 5 in an online manner. {Bold} text indicates the best performance.}
\label{tab:imagenetc}
\small
\centering
\setlength{\tabcolsep}{1mm} 
\begin{tabular}{l|l|ccccccccccccccc|c} 
\toprule
~ & Time & \multicolumn{15}{c|}{$t\xrightarrow{\makebox[\dimexpr 32\width][c]{\quad}}$} & ~ \\
\midrule
\multirow{4}{*}{Method} & \multirow{4}{*}{Venue} & \multirow{4}*{\rotatebox{75}{Gaussian}} & \multirow{4}*{\rotatebox{75}{shot}} & \multirow{4}*{\rotatebox{75}{impulse}} & \multirow{4}*{\rotatebox{75}{defocus}} & \multirow{4}*{\rotatebox{75}{glass}} & \multirow{4}*{\rotatebox{75}{motion}} & \multirow{4}*{\rotatebox{75}{zoom}} & \multirow{4}*{\rotatebox{75}{snow}} & \multirow{4}*{\rotatebox{75}{frost}} & \multirow{4}*{\rotatebox{75}{fog}} & \multirow{4}*{\rotatebox{75}{brightness}} & \multirow{4}*{\rotatebox{75}{contrast}} & \multirow{4}*{\rotatebox{75}{elastic}} & \multirow{4}*{\rotatebox{75}{pixelate}} & \multirow{4}*{\rotatebox{75}{jpeg}} & \multirow{4}*{{Mean$\downarrow$}} \\
~ & ~ & ~ & ~ & ~ & ~ & ~ & ~ & ~ & ~ & ~ & ~ & ~ & ~ & ~ & ~ & ~ \\
~ & ~ & ~ & ~ & ~ & ~ & ~ & ~ & ~ & ~ & ~ & ~ & ~ & ~ & ~ & ~ & ~ \\
~ & ~ & ~ & ~ & ~ & ~ & ~ & ~ & ~ & ~ & ~ & ~ & ~ & ~ & ~ & ~ & ~ \\
\midrule
Source & ICLR'21 & 62.6 & 64.2 & 61.9 & 70.2 & 75.3 & 68.2 & 68.7 & 63.5 & 55.5 & 49.9 & 31.7 & 78.1 & 60.5 & 47.5 & 46.9 & 60.3 \\
TENT & ICLR'21 & 60.3 & 58.5 & 53.5 & 70.7 & 71.3 & 64.0 & 66.7 & 58.6 & 49.9 & 43.5 & 30.9 & 67.5 & 55.8 & 45.2 & 44.1 & 56.0 \\
CoTTA & CVPR'22 & 61.9 & 61.6 & 57.9 & 68.6 & 71.0 & 63.8 & 65.0 & 58.1 & 48.7 & 43.7 & 29.8 & 71.9 & 49.3 & 41.4 & 39.7 & 55.5 \\
EATA & ICML'22 & 55.3 & 49.8 & 46.3 & 68.0 & 63.7 & 55.0 & 58.1 & 52.7 & 42.9 & 40.4 & 28.8 & 55.6 & 49.6 & 40.9 & 39.5 & 49.8 \\
RMT & CVPR'23 & 57.0 & 55.9 & 53.7 & 61.7 & 62.2 & 54.8 & 57.7 & 45.7 & 42.0 & 37.1 & 28.8 & 55.6 & 41.0 & 35.3 & 34.7 & 48.2 \\
CMAE & CVPR'24 & 48.2 & 46.6 & 47.0 & 64.1 & {62.9} & 50.7 & {62.9} & {59.1} & {43.2} & {34.2} & 28.5 & {79.9} & {65.8} & {41.3} & {39.7} & {51.6} \\
OBAO & ECCV'24 & 51.4 & 49.3 & 48.4 & 63.8 & {62.3} & 54.8 & {54.5} & {49.4} & {45.0} & {42.4} & 29.3 & {56.0} & {42.9} & {37.3} & {37.5} & {48.3} \\
DPCore & ICML'25 & 50.8 & 45.9 & 46.1 & 64.3 & 57.0 & 52.4 & {51.5} & {46.6} & {41.8} & {40.5} & 28.6 & {65.3} & {42.7} & {41.9} & {39.2} & 47.6 \\
\midrule
DDA & CVPR'23 & 43.0 & 48.0 & 46.6 & 72.3 & 66.5 & 67.6 & 67.2 & 63.9 & 54.4 & 70.4 & 36.6 & 89.3 & 50.3 & 41.6 & 43.1 & 57.4 \\
SDA & CVPR'25 & 41.4 & 47.4 & 45.7 & 70.7 & 64.3 & 65.0 & 65.6 & 61.5 & 54.3 & 69.9 & 35.0 & 89.3 & 47.5 & 39.0 & 40.3 & 55.8 \\
\midrule
Ours & Proposed & 48.9 & 45.9 & 45.6 & 54.9 & {56.5} & 43.4 & {54.9} & {40.7} & {39.8} & {34.7} & 29.2 & {43.7} & {43.2} & {42.2} & {37.5} & \textbf{44.1} \\
\bottomrule
\end{tabular}
\end{table*}

Through this multi-level design, the static knowledge base is dynamically transformed into a time-varying form that is highly aligned with the current target domain across visual appearance, feature statistics, and semantic representation. This process yields on-demand supervisory signals closely coupled with the instantaneous data distribution, thereby robustly supporting our forward-facilitation paradigm of continual adaptation.

\subsection{Optimization Objective}
Our synthetic knowledge base provides explicit semantic labels that, after dynamic style bridging, deliver precise supervision signals for current distribution adaptation. For notational simplicity, we denote stylized synthetic samples post multi-level bridging as $\tilde{x}_i^K$. We leverage the ground-truth labels of these transformed proxies $(\tilde{x}_i^K, y_i^K)$ in our time-varying knowledge base to supervise the predictions of the model $f_{\theta}$, which we term proxy-based cross-entropy ($\mathcal{L}_{PCE}$), is defined as:
\begin{equation}
    \mathcal{L}_{PCE} = -\sum_{c=1}^{C} y_{i,c}^K \log p_{i,c},
\end{equation}
where $p_i=\text{softmax}(f_{\theta}(\tilde{x}_i^K))$ represents the model's predicted probability distribution. This constitutes the primary supervision signal for our framework. Following prior methods~\cite{Dobler_2023_CVPR,zhu2024reshaping}, we sample a batch from the knowledge base with the same size as the target batch $B_t$ at each time step for loss computation. As the style of $\tilde{x}_i^K$ is aligned with the current target domain, $\mathcal{L}_{PCE}$ provides a direct and unbiased signal for transferring semantic knowledge.

To ensure stability during online adaptation and facilitate smooth knowledge transfer, we adopt a practical method widely established in teacher-student self-training~\cite{Dobler_2023_CVPR,zhu2024reshaping,ni2025maintaining} and compute the symmetric cross-entropy loss on target samples $B_t$, donated as $\mathcal{L}_{ST}$ (the detailed formulation is provided in the supplementary material). Finally, the total loss function is formulated as:
\begin{equation}
    \mathcal{L} = \mathcal{L}_{PCE}+\mathcal{L}_{SCL}+\mathcal{L}_{ST}.
\end{equation}

\section{Experiments}
In this section, we extensively evaluate the effectiveness of our proposed method under standard CTTA protocols~\cite{Wang_2022_CVPR,Dobler_2023_CVPR,zhu2024reshaping}. We first present the experimental settings and implementation details, and then provide a comprehensive comparative analysis of our method against previous state-of-the-art (SOTA) baselines. Furthermore, we construct thorough ablation studies toward an understanding of its working mechanisms. More detailed results and analyses are presented in the supplementary material. 

\subsection{Experimental Settings}
\paragraph{Benchmarks.} We construct extensive experiments on standard CTTA benchmarks: ImageNet-to-ImageNetC, CIFAR100-to-CIFAR100C, and CIFAR10-to-CIFAR10C. The original ImageNet and CIFAR datasets serve as source domains, while their corresponding corruption datasets~\cite{hendrycks2018benchmarking}, namely ImageNetC, CIFAR100C, and CIFAR10C, are utilized as target domains. Each corruption dataset comprises 15 distinct types of corruption at varying severity levels ranging from 1 to 5. Following standard task settings in~\cite{Dobler_2023_CVPR,zhu2024reshaping,liu2024continual}, we sequentially adapt the source model to these 15 target domains at the highest severity level 5. The entire adaptation process is conducted in a fully online manner, where the model receives no prior indication of domain shifts and must adapt on-the-fly.

\begin{table*}[htb]
\caption{Comparison results of standard CIFAR100-to-CIFAR100C and CIFAR10-to-CIFAR10C CTTA tasks. We report the mean classification error rate (\%, lower is better) across all 15 corrupted domains. All results are evaluated with the largest corruption severity level 5 in an online manner. {Bold} text indicates the best performance.}
\label{tab:CIFAR}
\centering
\setlength{\tabcolsep}{1mm} 
\small
\begin{tabular}{c|l|l|ccccccccccccccc|c} 
\toprule
~ & ~ & Time & \multicolumn{15}{c|}{$t\xrightarrow{\makebox[\dimexpr 32\width][c]{\quad}}$} & ~ \\
\midrule
~ & \multirow{4}{*}{Method} & \multirow{4}{*}{Venue} & \multirow{4}*{\rotatebox{75}{Gaussian}} & \multirow{4}*{\rotatebox{75}{shot}} & \multirow{4}*{\rotatebox{75}{impulse}} & \multirow{4}*{\rotatebox{75}{defocus}} & \multirow{4}*{\rotatebox{75}{glass}} & \multirow{4}*{\rotatebox{75}{motion}} & \multirow{4}*{\rotatebox{75}{zoom}} & \multirow{4}*{\rotatebox{75}{snow}} & \multirow{4}*{\rotatebox{75}{frost}} & \multirow{4}*{\rotatebox{75}{fog}} & \multirow{4}*{\rotatebox{75}{brightness}} & \multirow{4}*{\rotatebox{75}{contrast}} & \multirow{4}*{\rotatebox{75}{elastic}} & \multirow{4}*{\rotatebox{75}{pixelate}} & \multirow{4}*{\rotatebox{75}{jpeg}} & \multirow{4}*{{Mean$\downarrow$}} \\
~ & ~ & ~ & ~ & ~ & ~ & ~ & ~ & ~ & ~ & ~ & ~ & ~ & ~ & ~ & ~ & ~ & ~ \\
~ & ~ & ~ & ~ & ~ & ~ & ~ & ~ & ~ & ~ & ~ & ~ & ~ & ~ & ~ & ~ & ~ & ~ \\
~ & ~ & ~ & ~ & ~ & ~ & ~ & ~ & ~ & ~ & ~ & ~ & ~ & ~ & ~ & ~ & ~ & ~ \\
\midrule
\multirow{10}*{\rotatebox{90}{CIFAR100C}} & Source & ICLR'21 & 66.1 & 63.8 & 72.7 & 35.4 & 64.1 & 36.2 & 31.0 & 27.2 & 32.0 & 35.9 & 17.6 & 47.7 & 38.5 & 49.0 & 42.2 & 44.0 \\
~ & TENT & ICLR'21 & 65.9 & 62.9 & 71.2 & 35.0 & 64.3 & 35.6 & 30.2 & 27.7 & 32.3 & 34.9 & 17.5 & 44.8 & 38.0 & 50.1 & 41.9 & 43.5 \\
~ & CoTTA & CVPR'22 & 65.3 & 61.9 & 80.3 & 34.8 & 62.8 & 34.1 & 29.1 & 26.4 & 29.4 & 34.6 & 17.1 & 45.7 & 35.5 & 45.0 & 38.6 & 42.7 \\
~ & EATA & ICML'22 & 59.4 & 48.7 & 50.3 & 34.0 & 55.7 & 32.0 & 26.4 & 28.4 & 28.4 & 31.4 & 17.7 & 35.6 & 36.5 & 45.3 & 39.7 & 38.0 \\
~ & RMT & CVPR'23 & 61.2 & 54.7 & 58.1 & 31.1 & 53.4 & 28.2 & 24.3 & 23.6 & 24.5 & 26.5 & 17.2 & 23.9 & 24.9 & 25.8 & 28.5 & 33.7 \\
~ & CMAE & CVPR'24 & 54.3 & 47.3 & 61.2 & 34.0 & {75.5} & 31.6 & {32.0} & {25.2} & {28.1} & {27.8} & 16.3 & {25.6} & {36.5} & {37.0} & {39.6} & {38.2} \\
~ & OBAO & ECCV'24 & 56.3 & 50.8 & 57.6 & 31.2 & {54.3} & 29.2 & {24.2} & {23.5} & {24.0} & {27.2} & 17.5 & {26.3} & {26.9} & {29.1} & {30.5} & {33.9} \\
~ & REM & ICML'25 & 47.1 & 39.0 & 38.6 & 30.7 & {53.5} & 31.3 & {28.3} & {31.9} & {29.3} & {32.6} & 24.5 & {28.4} & {44.9} & {35.1} & {50.4} & {36.4} \\
~ & SDA & CVPR'25 & 53.1 & 51.0 & 70.3 & 38.2 & 66.6 & 38.8 & 32.8 & 28.3 & 32.6 & 38.8 & 19.1 & 40.2 & 41.2 & 48.1 & 44.9 & 42.9 \\
~ & Ours & Proposed & 43.8 & 35.7 & 35.6 & 27.2 & {50.0} & 25.0 & {22.8} & {24.1} & {23.7} & {23.8} & 17.9 & {20.1} & {31.6} & {28.1} & {37.8} & \textbf{29.8} \\
\midrule
\multirow{10}*{\rotatebox{90}{CIFAR10C}} & Source & ICLR'21 & 37.3 & 34.0 & 44.5 & 12.3 & 26.4 & 14.0 & 11.1 & 7.1 & 9.5 & 13.1 & 3.9 & 28.4 & 13.3 & 18.1 & 15.3 & 19.2 \\
~ & TENT & ICLR'21 & 30.8 & 21.9 & 29.4 & 14.0 & 21.9 & 15.0 & 10.1 & 7.7 & 8.1 & 12.6 & 4.2 & 19.3 & 12.8 & 18.3 & 14.7 & 16.1 \\
~ & CoTTA & CVPR'22 & 49.3 & 45.9 & 55.2 & 12.3 & 25.7 & 12.3 & 9.0 & 7.1 & 7.9 & 12.0 & 3.8 & 25.6 & 11.6 & 15.2 & 12.9 & 20.4 \\
~ & EATA & ICML'22 & 26.3 & 16.4 & 21.3 & 11.2 & 19.6 & 12.5 & 8.2 & 6.8 & 6.8 & 9.4 & 4.1 & 12.8 & 11.8 & 13.6 & 13.6 & 13.0 \\
~ & RMT & CVPR'23 & 31.2 & 25.2 & 23.7 & 8.8 & 17.9 & 8.2 & 5.7 & 5.4 & 6.2 & 7.0 & 3.5 & 5.1 & 6.5 & 5.5 & 7.7 & 11.2 \\
~ & CMAE & CVPR'24 & 27.6 & 21.5 & 23.4 & 8.7 & {23.8} & 9.5 & {6.6} & {6.2} & {7.2} & {8.3} & 3.4 & {8.8} & {11.6} & {9.6} & {12.6} & {12.6} \\
~ & OBAO & ECCV'24 & 33.1 & 28.2 & 28.9 & 9.0 & {18.8} & 8.1 & {5.8} & {5.3} & {6.0} & {6.7} & 3.6 & {5.0} & {6.3} & {6.4} & {8.2} & {12.0} \\
~ & REM & ICML'25 & 17.8 & 13.6 & 14.1 & 8.7 & {18.3} & 10.0 & {6.8} & {8.4} & {6.7} & {8.5} & 5.1 & {7.5} & {14.5} & {10.5} & {16.4} & {11.1} \\
~ & SDA & CVPR'25 & 23.8 & 21.7 & 43.1 & 11.9 & 27.6 & 13.7 & {10.9} & {7.4} & {9.8} & {13.1} & 3.8 & 15.3 & 14.1 & 19.0 & 16.5 & 16.8 \\
~ & Ours & Proposed & 17.0 & 12.0 & 11.3 & 7.9 & {18.4} & 7.1 & {5.6} & {6.3} & {6.2} & {5.6} & 3.8 & {4.9} & {9.9} & {8.8} & {12.0} & \textbf{9.1} \\
\bottomrule
\end{tabular}
\end{table*}

\paragraph {Comparison Methods.}
We compare our approach with diverse SOTA CTTA methods. These encompass single model-based methods such as Tent~\cite{wang2021tent}, EATA~\cite{niu2022efficient}, DPCore~\cite{Zhang2025DPCoreDP} and REM~\cite{han2025ranked}, as well as teacher-student framework-based methods, comprising CoTTA~\cite{Wang_2022_CVPR}, RMT~\cite{Dobler_2023_CVPR}, CMAE~\cite{liu2024continual}, and OBAO~\cite{zhu2024reshaping}. We also consider two diffusion-based methods, DDA~\cite{gao2023back} and SDA~\cite{guo2025everything}, which denoise corrupted test-time samples via generative modeling. Among them, we highlight OBAO and SDA as our primary competitors: OBAO dynamically aggregates high-confidence target samples during test time, while SDA projects target samples back to the synthetic domain. In contrast, our approach dynamically injects synthetic knowledge into an arbitrary current domain through a multi-level bridging mechanism, enabling more flexible and effective adaptation to evolving distributions. 

\paragraph {Implementation Details.}
Following prior studies~\cite{Dobler_2023_CVPR,liu2024continual,Zhang2025DPCoreDP}, we adopt the ViT-B/16 backbone~\cite{dosovitskiyimage}, initialized with ImageNet-1K pre-trained weights at a resolution of 224×224, as provided by the timm library~\cite{rw2019timm}. For the CIFAR-based experiments, we further fine-tune the model on the respective source domains to obtain task-specific source models. To ensure fairness, all methods are initialized with the same pre-trained weights for each dataset, and the batch size is fixed to 50. An Adam optimizer with a learning rate of 1e-5 is used to optimize the model. The loss terms are assigned equal weights, which was found to be sufficient for stable optimization across benchmarks. For synthetic knowledge construction, we employ Stable Diffusion 1.5~\cite{rombach2022high} to generate few-shot class-conditional samples per category. Importantly, our approach does not require any interaction with the diffusion model during test time. All ablation studies are conducted on ImageNet-to-ImageNetC unless otherwise specified. More details for reproducibility and training regimes of each method are provided in the supplementary material.

\subsection{Comparison Results}
\paragraph{Results on ImageNet-to-ImageNetC.} Given a source model pre-trained on ImageNet-1K, we conduct CTTA sequentially across all 15 corruption domains in ImageNetC. The model continually adapts to each target domain over time, and we report the average classification error across all domains. As shown in~\cref{tab:imagenetc}, directly evaluating the source model under corrupted conditions yields an average error rate up to 60.3\%, underscoring the necessity of test-time adaptation. Various baseline methods have achieved performance improvements by utilizing unlabeled test samples for model adaptation. DDA applies denoising via diffusion models to counter distribution shifts at the input level, and SDA further fine-tunes the pre-trained model before adaptation to align with the synthetic domain. However, their reliance on computationally intensive test-time denoising makes them impractical for real-world deployment.

In our method, we explicitly incorporate semantic information from the synthetic domain into the test process through a dynamic style bridging mechanism, directly addressing the central challenge of limited supervision under the CTTA setting. Our approach outperforms all previous methods, significantly reducing the average classification error from 60.3\% of the source model to 44.1\%, achieving a relative 7.3\% improvement over the existing SOTA method, DPCore. While image-level adaptation methods perform well on specific corruptions (\eg, Gaussian, Shot, and Impulse noise), they struggle under broader test conditions, exposing the limitations of denoising-based strategies. By contrast, our method achieves a relative improvement of up to 20.9\% over SDA and consistently delivers strong performance across a wide range of corruptions. This highlights not only its effectiveness in mitigating catastrophic forgetting but also its enhanced robustness and adaptability to diverse and unpredictable test-time conditions.

\paragraph{Results on CIFAR Benchmarks.} To further validate the effectiveness of our method, we evaluate it on CIFAR100-to-CIFAR100C and CIFAR10-to-CIFAR10C benchmarks, which differ in the number of categories and task complexity. As summarized in~\cref{tab:CIFAR}, our method consistently outperforms all competing approaches, achieving average error rates of 29.8\% and 9.1\% on the respective tasks. Due to the low resolution of test samples, methods that rely on input-level denoising during inference tend to underperform. In contrast, our method integrates explicit semantic information from the synthetic domain into CTTA and dynamically evolves this information through the proposed multi-level bridging mechanism, yielding substantial performance gains. For instance, on CIFAR100C, our method achieves a relative mean error rate reduction of 30.5\% compared to SDA. The consistent improvements across datasets of varying complexity highlight both the effectiveness and universality of our proposed framework.

\subsection{Ablation Studies and Analysis}
\begin{table}[t]
\caption{Ablation experiments on the standard ImageNet-to-ImageNetC CTTA task. Each component of our framework is progressively activated to analyze its contribution.}
\label{tab:ab}
\centering
\small
\setlength{\tabcolsep}{1mm} 
\begin{tabular}{c|cccc|c} 
\toprule
~ & $\mathcal{L}_{PCE}$ & Input & Statistic & $\mathcal{L}_{SCL}$ & Mean$\downarrow$ \\
\midrule
$Ex_1$ & - & - & - & - & 50.0 \\
$Ex_2$ & \checkmark & - & - & - & 47.4 \\
$Ex_3$ & \checkmark & \checkmark & - & - & 45.8 \\
$Ex_4$ & \checkmark & - & \checkmark & - & 45.4 \\
$Ex_5$ & \checkmark & - & - & \checkmark & 46.5 \\
$Ex_6$ & \checkmark & \checkmark & \checkmark & - & 44.6 \\
$Ex_7$ & \checkmark & \checkmark & \checkmark & \checkmark & 44.1 \\
\bottomrule
\end{tabular}

\end{table}

\paragraph{Effect of Each Component.}
We conduct a comprehensive ablation study to validate the effectiveness of each component in our framework, as summarized in~\cref{tab:ab}. $Ex_1$ denotes the baseline performance using only self-training loss, yielding a relatively high average error rate due to the lack of reliable supervision. In $Ex_2$, we introduce our synthetic knowledge base without any style bridging. This provides explicit semantic information into the CTTA process and clearly improves performance, but the synthetic exemplars still carry noticeable generative bias. From $Ex_3$ to $Ex_7$, we substantiate the efficacy of the proposed multi-level bridging mechanism. Through progressive transformations across input, statistics, and representation levels, we customize the static synthetic domain knowledge to the current distribution, precisely delivering the requisite supervision signals for the model's adaptation process. Ultimately, $Ex_7$ shows significant overall improvement, demonstrating that our proposed components are synergistically complementary and collectively enhance adaptation performance.

\begin{figure}[t]
\centering
\includegraphics[width=0.90\columnwidth]{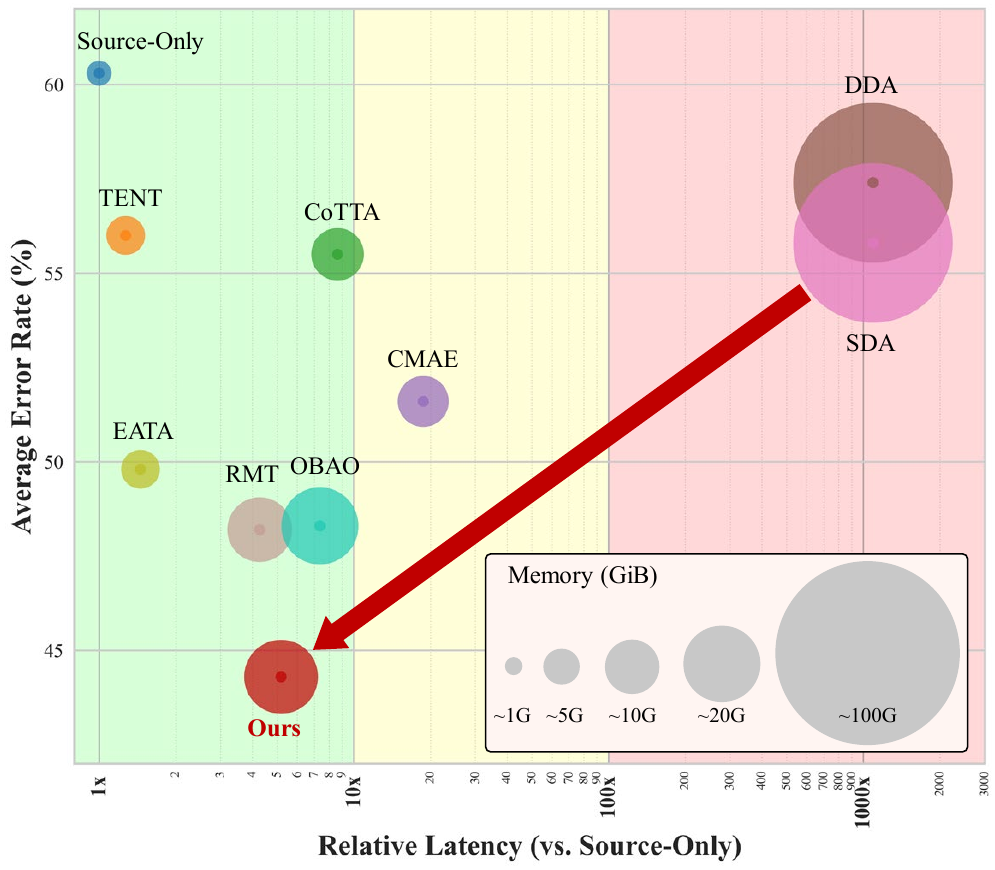} 
\caption{Comparison between different methods in terms of average error rate (\%), latency relative to the source model, and GPU memory overhead on the standard ImageNet-to-ImageNetC task.}
\label{fig:resources}
\end{figure}

\begin{table}[t]
\caption{Efficiency comparison. We compare the GPU memory (GiB), relative latency (Source=1.0), and mean error rate (\%). Results are reported on CIFAR100-to-CIFAR100C.}
\label{tab:Ablation_efficiency_100c}
\centering
\small
\setlength{\tabcolsep}{1mm} 
\begin{tabular}{lccc}
\toprule
Method & GPU Mem$\downarrow$ & Latency$\downarrow$ & Mean error$\downarrow$ \\
\midrule
Source           & 2.1  & 1.0 & 44.0 \\
TENT             & 5.3  & 2.0 & 43.5 \\
CoTTA            & 9.8  & 22.0  & 42.7 \\
CMAE             & 9.5  & 36.1 & 38.2 \\
OBAO             & 21.6  & 10.3 & 33.9 \\
SDA              & 95.8 & 1719.2 &  42.9 \\
Ours             & 19.9 & 9.0 &  \textbf{29.8} \\
\bottomrule
\end{tabular}
\end{table}

\paragraph{Efficiency Comparison.}
~\cref{fig:resources} visualizes the resource consumption of our method against competing approaches. 
Our approach substantially improves the effectiveness and efficiency of the utilization of synthetic knowledge.
Additionally, we provide detailed values on another benchmark in~\cref{tab:Ablation_efficiency_100c}. For practical deployment in CTTA, efficiency metrics such as GPU memory usage and inference latency constitute important considerations. Lightweight methods like TENT exhibit low resource usage but suffer from limited performance. Diffusion-based approaches such as SDA incur excessive computational demands, rendering them impractical for real-time online adaptation. In contrast, our approach achieves \ZLChange{a favorable balance.}

\begin{figure}[t]
\centering
\includegraphics[width=0.8\columnwidth]{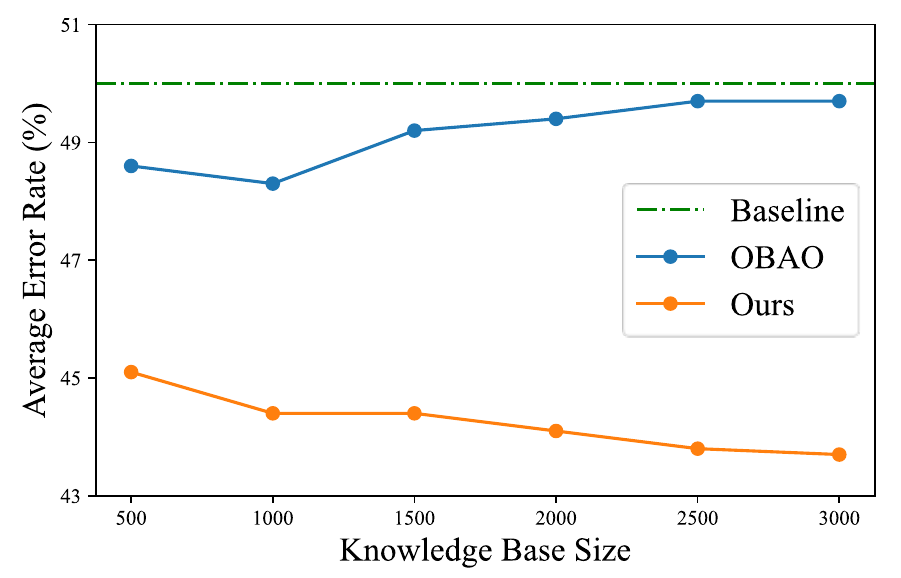} 
\caption{Ablation study on the size of the knowledge base.}
\label{fig:nums}
\end{figure}
\paragraph{Effect of Knowledge Base Size.} \cref{fig:nums} illustrates the ablation study examining the size of our constructed synthetic knowledge base, compared against OBAO, which similarly employs a knowledge base. Note that OBAO dynamically buffers low-entropy samples in the evolving data stream, where the noise problem of pseudo-labeling inevitably exacerbates as the buffer capacity gets larger. In contrast, our framework directly integrates explicit semantic information of the synthetic domain. Remarkably, by incorporating merely 1 synthetic example per class, our method achieves excellent performance and exhibits insensitivity when the count exceeds 2. Based on this observation, we set the size of the synthetic knowledge base to 2000. Through our multi-level bridging mechanism, the requisite optimization signals for model adaptation can be continuously supplied, while significantly outperforming the comparison method and imposing only negligible storage overhead.

\begin{table}
  \caption{Quantitative results using different generative models.}
  \setlength{\tabcolsep}{1mm} 
  \small
  \label{tab:generators}
  \centering
  \begin{tabular}{cccc}
    \toprule
    Generator & BigGAN & SD 1.5 & SD 3.0\\
    \midrule
    Ours & 45.0 & 44.1 & 43.8 \\
    \bottomrule
  \end{tabular}
\end{table}

\begin{figure}[t]
\centering
\includegraphics[width=0.8\columnwidth]{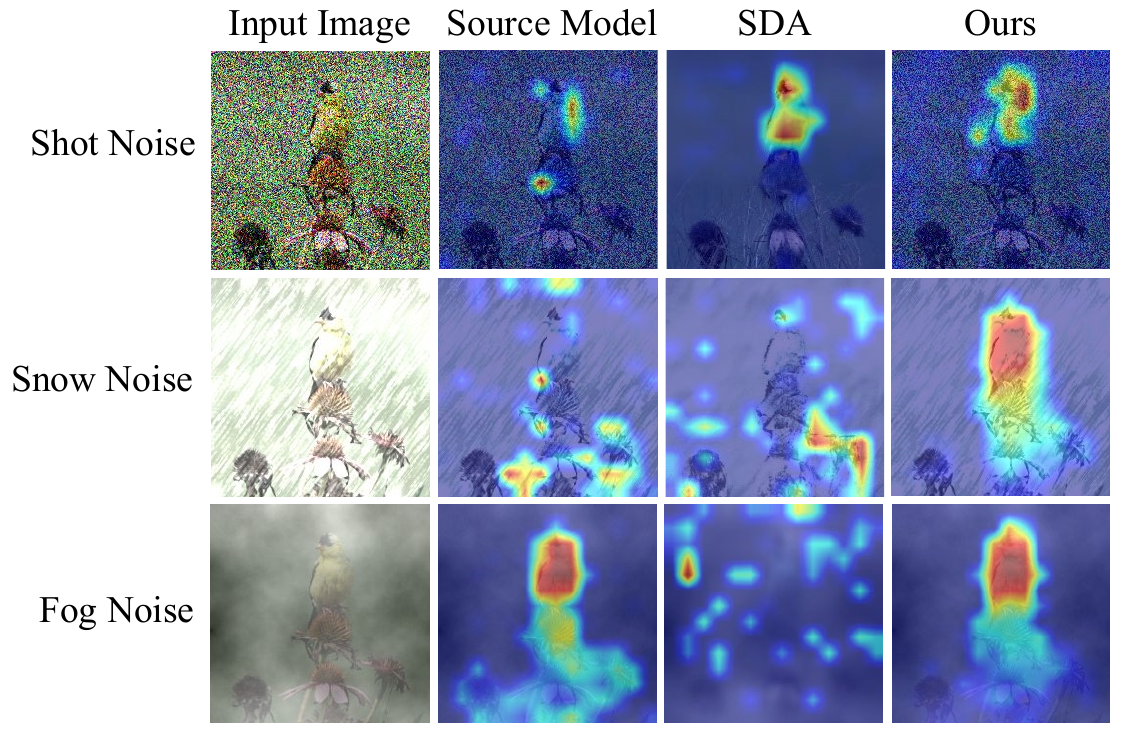} 
\caption{Grad CAM visualization. Each row corresponds to the results for
shot, snow, and fog noise, respectively.}
\label{fig:grad-cam}
\end{figure}

\paragraph{Robustness to Generators.}
We evaluate the influence of different generative models on the proposed framework. Specifically, BigGAN~\cite{brock2018large}, Stable Diffusion 1.5~\cite{rombach2022high}, and Stable Diffusion 3.0~\cite{esser2024scaling} are employed to construct knowledge bases of identical scale. As shown in~\cref{tab:generators}, our framework demonstrates remarkable robustness across various generators. Moreover, the performance trend suggests that the framework is well-positioned to leverage the ongoing advances in generative modeling techniques. Visualizations of samples produced by each generator are available in the supplementary material.

\paragraph{Class Attention Map Visualization.}
~\cref{fig:grad-cam} illustrates Grad-CAM visualization results that reveal the salient regions influencing model predictions. The source model exhibits substantial variations in feature attention across domains, indicating pronounced susceptibility to domain bias. While the comparative method SDA mitigates noise by projecting test samples onto the synthetic domain, its efficacy remains limited to specific corruption scenarios and degrades substantially under broader environmental perturbations. In contrast, our method enables the model to learn discriminative representations across conditions and consistently highlights foreground objects with higher activation values, demonstrating superior robustness and adaptability.

\section{Conclusion}
In this paper, we re-examine CTTA from the perspective of supervision reliability. Prior methods typically rely on rigid backward alignment from incoming data to source-domain surrogates, leading to noisy self-training signals. In contrast, we propose a novel forward-facilitation paradigm. This approach begins by generating a compact set of representative samples according to the on-demand supervision labels, which then co-evolve with the test distribution via a multi-level bridging mechanism. The key benefit is consistently reliable supervision signals that mitigate the problem of noise accumulation. Extensive experiments on standard benchmarks validate the effectiveness and robustness of our proposed framework.

{
    \small
    \bibliographystyle{ieeenat_fullname}
    \bibliography{main}
}

\maketitlesupplementary

\appendix 

\setcounter{table}{5} 
\setcounter{figure}{5} 

In the supplementary material, we provide more analysis and experiments to enhance the understanding of our work. In~\cref{method_implementations}, we provide comprehensive information about different CTTA baselines and implementations involved in our main paper. \cref{sec:supp_fig1} provides a detailed illustration of symbols used in paradigm comparison. \cref{L_ST} shows details and compatibility of our framework across self-training objectives. In~\cref{additional_results}, we provide additional experimental results to further validate the generalizability of our method. Finally, we present visualizations of synthetic images in~\cref{Visualization_Synthetic_Data}.

\section{Details of Comparison Methods}\label{method_implementations}
In this section, we provide details of the CTTA methods included in our comparisons of the main paper.

\noindent\textbf{TENT}\footnote{https://github.com/DequanWang/tent}~\cite{wang2021tent} makes all affine parameters of normalization layers trainable through the entropy minimization loss. We follow all hyperparameters that are set in TENT unless it does not provide.

\noindent\textbf{CoTTA}\footnote{https://github.com/qinenergy/cotta}~\cite{Wang_2022_CVPR} represents the first method to perform TTA on continually changing domains. For its augmentation-based consistency maximization and stochastic parameter restoration, we maintain identical hyperparameters as specified in CoTTA. The trainable parameters encompass all parameters within ViT-Base.

\noindent\textbf{EATA}\footnote{https://github.com/mr-eggplant/EATA}~\cite{niu2022efficient} selectively performs test-time optimization
with relatively reliable and non-redundant samples. For its thresholds of entropy filter and cosine similarities, we configure the same hyperparameters. Similar to TENT, the trainable parameters comprise all affine parameters of normalization layers.

\noindent\textbf{RMT}\footnote{https://github.com/mariodoebler/test-time-adaptation}~\cite{Dobler_2023_CVPR} adopts a mean-teacher framework with a symmetric cross-entropy loss, combined with contrastive learning to align with source representative features. For its model optimization and updating process, we configure the same hyperparameters. The trainable parameters are all the parameters in ViT-Base.

\noindent\textbf{CMAE}\footnote{https://github.com/RanXu2000/continual-mae}~\cite{liu2024continual} propose adaptive distribution masked autoencoders to enhance the extraction of target domain knowledge while mitigating the accumulation of distribution shifts. We follow the same uncertain calculation method and masking strategy.

\noindent\textbf{OBAO}\footnote{https://github.com/z1358/OBAO}~\cite{zhu2024reshaping} dynamically identifies and aggregates
high-confidence target samples during test time, combined with a class relation preservation constraint to organize these samples. For its dynamical buffer, we follow the same hyperparameters.

\noindent\textbf{DPCore}\footnote{https://github.com/yunbeizhang/DPCore}~\cite{Zhang2025DPCoreDP} utilizes a prompt coreset which dynamically manages domain knowledge to match the statistical prior in source domain. The trainable parameters are the introduced prompts. We follow all hyperparameters unless it does not provide.

\noindent\textbf{REM}\footnote{https://github.com/pilsHan/rem}~\cite{han2025ranked} proposes ranked entropy minimization to mitigate
the stability problem. A progressive masking strategy is introduced to explicitly structure the prediction difficulty. We follow the same masking strategy to build the mask chain.

\noindent\textbf{DDA}\footnote{https://github.com/shiyegao/DDA}~\cite{gao2023back} projects all test inputs
toward the source domain with a generative diffusion model. It deals with distribution shifts at the input level. We follow all hyperparameters and use the same diffusion model.

\noindent\textbf{SDA}\footnote{https://github.com/SHI-Labs/Diffusion-Driven-Test-Time-Adaptation-via-Synthetic-Domain-Alignment}~\cite{guo2025everything} is similar to DDA in terms of overall process. However, it further fine-tunes the source model in the synthesis domain. We utilize the same stable diffusion model~\cite{rombach2022high} of our method to generate training samples, and further refine them via an unconditional diffusion model.

\begin{table*}[ht]
\centering
\caption{Detailed explanation of graphical symbols used in paradigm comparison.}
\label{tab:fig1_symbols}
\small
\begin{tabular}{c|l|l}
\toprule
\textbf{Sub-figure} & \textbf{Symbol/Element} & \textbf{Technical Interpretation} \\
\midrule
\multirow{4}{*}{(b)} & Network icon & Parameter regularization (constraints or recovery on source weights) \\
 & Star ($\star$) & Source-domain class prototypes (feature centroids) used for representation alignment \\
 & $\{\mu^s, \sigma^s\}$ & Source feature statistics (mean and variance) / BN statistics \\
 & Align arrow & Backward alignment that constrains the target model using source priors \\
\midrule
\multirow{2}{*}{(c)} & $x^t \rightarrow x^K$ & Projecting a target input back to a static synthetic domain \\
 & Denoise arrow & The process of removing ``domain shift" as noise \\
\midrule
\multirow{2}{*}{(d)} & $\{x^K, y^K\}$ & Explicit synthetic image-label pairs (Knowledge Base) \\
 & Co-evolve arrow & Actively injecting target styles into synthetic knowledge (Bridging) \\
\bottomrule
\end{tabular}
\end{table*}

\subsection{Experimental protocols}\label{protocols_implementations}
The pretrained model is ViT-B/16~\cite{dosovitskiyimage}, trained on the ImageNet-1K training set at a resolution of 224×224, with weights directly obtained from the \texttt{timm} repository. The same data preprocessing configuration is adopted to maintain consistency. For the CIFAR-based experiments, we further fine-tune the model on the respective source domains to obtain task-specific source models. To ensure fairness, all methods are initialized with the same pre-trained weights for each dataset, and the batch size is fixed to 50. We utilize official implementations of the method where available. Note that some methods only provide implementations on convolutional neural networks, at which point we prioritize the reproduction with the help of existing methods~\cite{liu2024vida,han2025ranked}. If not, we reproduce it ourselves using the hyperparameters reported in the original paper. Optimization is performed using Adam with $(\beta_1, \beta_2) = (0.9, 0.999)$ and the learning rate is chosen from $\{1e-3, 1e-4, 1e-5\}$ to achieve the proper magnitude.

\subsection{More Implementation Details}\label{implementation_details}
Our approach follows the experimental protocols outlined in~\cref{protocols_implementations} to ensure consistency and comparability. For the construction of our synthetic knowledge base, we utilize Stable Diffusion 1.5~\cite{rombach2022high} with 50 denoising steps, generating 2000 synthetic images per CTTA task. Importantly, our approach does not require any interaction with the diffusion model during test time. Following the prior method~\cite{zhu2024reshaping}, we sample a batch from the knowledge base with the same size as the target batch $B_t$ at each time step for loss computation. An Adam optimizer with a learning rate of 1e-5 is used to optimize the model. For the teacher-student framework involved in the self-training loss, since our forward-facilitation paradigm can provide the necessary supervision information for model adaptation, we use a larger momentum of 0.9 to update the teacher model through the exponential moving average. All experiments of our method in the main paper during test time are conducted using an NVIDIA RTX4090 GPU.

\section{Detailed Illustration of Symbols in Fig.~1}
\label{sec:supp_fig1}
Fig.~1 in the main paper illustrates the conceptual distinctions between the prevailing backward-alignment paradigm and our proposed forward-facilitation paradigm. To ensure a precise understanding of the visual schematic, we provide a detailed explanation of the key graphical symbols appearing in sub-figures (b)-(d). We summarize these elements in~\cref{tab:fig1_symbols}. The provided definitions bridge the conceptual illustration with the methodological details in the main text, enabling readers to interpret the figure accurately and without ambiguity.

\begin{table}[ht]
  \caption{Compatibility with different self-training objectives. We report the average error rate (\%, lower is better) of the baseline and our method. Consistent improvements are yielded in both settings.}
  \small
  \label{tab:self_training_variants}
  \centering
  \begin{tabular}{cccc}
    \toprule
    Self-training & Baseline & Ours & $\Delta \downarrow$ \\
    \midrule
    Entropy minimization & 49.8 & {44.8} & {-5.0} \\
    Teacher-student & 50.0 & \textbf{44.1} & \textbf{-5.9} \\
    \bottomrule
  \end{tabular}
\end{table}

\begin{table*}[!th]
\caption{Classification error rate (\%, lower is better) for the standard CTTA task on ImageNet-to-ImageNetC. We follow the experimental protocol established by DPCore. {Bold} text indicates the best.}
\label{tab:imagenetc_21k_ft_1k}
\small
\centering
\setlength{\tabcolsep}{1mm} 
\begin{tabular}{l|l|ccccccccccccccc|c} 
\toprule
~ & Time & \multicolumn{15}{c|}{$t\xrightarrow{\makebox[\dimexpr 32\width][c]{\quad}}$} & ~ \\
\midrule
\multirow{4}{*}{Method} & \multirow{4}{*}{Venue} & \multirow{4}*{\rotatebox{75}{Gaussian}} & \multirow{4}*{\rotatebox{75}{shot}} & \multirow{4}*{\rotatebox{75}{impulse}} & \multirow{4}*{\rotatebox{75}{defocus}} & \multirow{4}*{\rotatebox{75}{glass}} & \multirow{4}*{\rotatebox{75}{motion}} & \multirow{4}*{\rotatebox{75}{zoom}} & \multirow{4}*{\rotatebox{75}{snow}} & \multirow{4}*{\rotatebox{75}{frost}} & \multirow{4}*{\rotatebox{75}{fog}} & \multirow{4}*{\rotatebox{75}{brightness}} & \multirow{4}*{\rotatebox{75}{contrast}} & \multirow{4}*{\rotatebox{75}{elastic}} & \multirow{4}*{\rotatebox{75}{pixelate}} & \multirow{4}*{\rotatebox{75}{jpeg}} & \multirow{4}*{{Mean$\downarrow$}} \\
~ & ~ & ~ & ~ & ~ & ~ & ~ & ~ & ~ & ~ & ~ & ~ & ~ & ~ & ~ & ~ & ~ \\
~ & ~ & ~ & ~ & ~ & ~ & ~ & ~ & ~ & ~ & ~ & ~ & ~ & ~ & ~ & ~ & ~ \\
~ & ~ & ~ & ~ & ~ & ~ & ~ & ~ & ~ & ~ & ~ & ~ & ~ & ~ & ~ & ~ & ~ \\
\midrule
Source & ICLR'21 & 53.0 & 51.8 & 52.1 & 68.5 & 78.8 & 58.5 & 63.3 & 49.9 & 54.2 & 57.7 & 26.4 & 91.4 & 57.5 & 38.0 & 36.2 & 55.8 \\
TENT & ICLR'21 & 52.2 & 48.9 & 49.2 & 65.8 & 73.0 & 54.5 & 58.4 & 44.0 & 47.7 & 50.3 & 23.9 & 72.8 & 55.7 & 34.4 & 33.9 & 51.0 \\
CoTTA & CVPR'22 & 52.9 & 51.6 & 51.4 & 68.3 & 78.1 & 67.1 & 62.0 & 48.2 & 52.7 & 55.3 & 25.9 & 90.0 & 56.4 & 36.4 & 35.2 & 54.8 \\
VDP & AAAI'23 & 52.7 & 51.6 & 50.1 & 58.1 & 70.2 & 56.1 & 58.1 & 42.1 & 46.1 & 45.8 & 23.6 & 70.4 & 54.9 & 34.5 & 36.1 & 50.0 \\
EcoTTA & CVPR'23 & 48.1 & 45.6 & 46.3 & 56.5 & 67.1 & 50.4 & 57.1 & 41.3 & 44.5 & 43.8 & 24.1 & 71.6 & 54.8 & 34.1 & 34.8 & 48.0 \\
CMAE & CVPR'24 & 46.3 & 41.9 & 42.5 & 51.4 & {54.9} & 43.3 & {40.7} & {34.2} & {35.8} & {64.3} & 23.4 & {60.3} & {37.5} & {29.2} & {31.4} & {42.5} \\
DPCore & ICML'25 & 42.2 & 38.7 & 39.3 & 47.2 & 51.4 & 47.7 & {46.9} & {39.3} & {36.9} & {37.4} & 22.0 & {44.4} & {45.1} & {30.9} & {29.6} & 39.9 \\
PAID & NeurIPS'25 & 48.8 & 43.7 & 44.4 & 49.4 & 49.6 & 47.3 & 44.2 & 37.5 & 39.4 & 42.1 & 25.2 & 50.0 & 39.3 & 35.5 & 36.5 & 42.2 \\
Ours & Proposed & 40.0 & 38.0 & 38.3 & 41.9 & {52.4} & 35.2 & {43.8} & {31.5} & {34.7} & {28.1} & 24.6 & {33.2} & {37.5} & {31.6} & {30.6} & \textbf{36.1} \\
\bottomrule
\end{tabular}
\end{table*}

\begin{table*}[!thb]
\caption{Comparison results of standard CIFAR100-to-CIFAR100C and CIFAR10-to-CIFAR10C CTTA tasks. We report the mean classification error rate (\%, lower is better) across all 15 corrupted domains. All results are evaluated with the largest corruption severity level 5 in an online manner. We follow the experimental protocol established by DPCore. {Bold} text indicates the best.}
\label{tab:CIFAR_384}
\centering
\setlength{\tabcolsep}{1mm} 
\small
\begin{tabular}{c|l|l|ccccccccccccccc|c} 
\toprule
~ & ~ & Time & \multicolumn{15}{c|}{$t\xrightarrow{\makebox[\dimexpr 32\width][c]{\quad}}$} & ~ \\
\midrule
~ & \multirow{4}{*}{Method} & \multirow{4}{*}{Venue} & \multirow{4}*{\rotatebox{75}{Gaussian}} & \multirow{4}*{\rotatebox{75}{shot}} & \multirow{4}*{\rotatebox{75}{impulse}} & \multirow{4}*{\rotatebox{75}{defocus}} & \multirow{4}*{\rotatebox{75}{glass}} & \multirow{4}*{\rotatebox{75}{motion}} & \multirow{4}*{\rotatebox{75}{zoom}} & \multirow{4}*{\rotatebox{75}{snow}} & \multirow{4}*{\rotatebox{75}{frost}} & \multirow{4}*{\rotatebox{75}{fog}} & \multirow{4}*{\rotatebox{75}{brightness}} & \multirow{4}*{\rotatebox{75}{contrast}} & \multirow{4}*{\rotatebox{75}{elastic}} & \multirow{4}*{\rotatebox{75}{pixelate}} & \multirow{4}*{\rotatebox{75}{jpeg}} & \multirow{4}*{{Mean$\downarrow$}} \\
~ & ~ & ~ & ~ & ~ & ~ & ~ & ~ & ~ & ~ & ~ & ~ & ~ & ~ & ~ & ~ & ~ & ~ \\
~ & ~ & ~ & ~ & ~ & ~ & ~ & ~ & ~ & ~ & ~ & ~ & ~ & ~ & ~ & ~ & ~ & ~ \\
~ & ~ & ~ & ~ & ~ & ~ & ~ & ~ & ~ & ~ & ~ & ~ & ~ & ~ & ~ & ~ & ~ & ~ \\
\midrule
\multirow{9}*{\rotatebox{90}{CIFAR100C}} & Source & ICLR'21 & 55.0 & 51.5 & 26.9 & 24.0 & 60.5 & 29.0 & 21.4 & 21.1 & 25.0 & 35.2 & 11.8 & 34.8 & 43.2 & 56.0 & 35.9 & 35.4 \\
~ & TENT & ICLR'21 & 53.0 & 47.0 & 24.6 & 22.3 & 58.5 & 26.5 & 19.0 & 21.0 & 23.0 & 30.1 & 11.8 & 25.2 & 39.0 & 47.1 & 33.3 & 32.1 \\
~ & CoTTA & CVPR'22 & 55.0 & 51.3 & 25.8 & 24.1 & 59.2 & 28.9 & 21.4 & 21.0 & 24.7 & 34.9 & 11.7 & 31.7 & 40.4 & 55.7 & 35.6 & 34.8 \\
~ & VDP & AAAI'23 & 54.8 & 51.2 & 25.6 & 24.2 & 59.1 & 28.8 & 21.2 & 20.5 & 23.3 & 33.8 & 7.5 & 11.7 & 32.0 & 51.7 & 35.2 & 32.0 \\
~ & ViDA & ICLR'24 & 50.1 & 40.7 & 22.0 & 21.2 & 45.2 & 21.6 & 16.5 & 17.9 & 16.6 & 25.6 & 11.5 & 29.0 & 29.6 & 34.7 & 27.1 & 27.3 \\
~ & CMAE & CVPR'24 & 48.6 & 30.7 & 18.5 & 21.3 & 38.4 & 22.2 & 17.5 & 19.3 & 18.0 & 24.8 & 13.1 & 27.8 & 31.4 & 35.5 & 29.5 & 26.4 \\
~ & DPCore & ICML'25 & 48.2 & 40.2 & 21.3 & 20.2 & 44.1 & 21.1 & 16.2 & 18.1 & 15.2 & 22.3 & 9.4 & 13.2 & 28.6 & 32.8 & 25.5 & 25.1 \\
~ & PAID & NeurIPS'25 & 40.7 & 31.9 & 20.4 & 19.8 & 35.9 & 23.0 & 16.3 & 20.5 & 18.2 & 25.3 & 12.6 & 19.8 & 29.4 & 28.2 & 31.3 & 24.9 \\
~ & Ours & Proposed & 35.2 & 29.8 & 19.4 & 19.1 & {36.0} & 19.5 & {17.2} & {15.8} & {16.7} & {18.6} & 11.6 & {13.0} & {27.6} & {25.0} & {27.9} & \textbf{22.2} \\
\midrule
\multirow{9}*{\rotatebox{90}{CIFAR10C}} & Source & ICLR'21 & 60.1 & 53.2 & 38.3 & 19.9 & 35.5 & 22.6 & 18.6 & 12.1 & 12.7 & 22.8 &  5.3 & 49.7 & 23.6 & 24.7 & 23.1 & 28.2 \\
~ & TENT & ICLR'21 & 57.7 & 56.3 & 29.4 & 16.2 & 35.3 & 16.2 & 12.4 & 11.0 & 11.6 & 14.9 & 4.7 & 22.5 & 15.9 & 29.1 & 19.5 & 23.5 \\
~ & CoTTA & CVPR'22 & 58.7 & 51.3 & 33.0 & 20.1 & 34.8 & 20.0 & 15.2 & 11.1 & 11.3 & 18.5 & 4.0 & 34.7 & 18.8 & 19.0 & 17.9 & 24.6 \\
~ & VDP & AAAI'23 & 57.5 & 49.5 & 31.7 & 21.3 & 35.1 & 19.6 & 15.1 & 10.8 & 10.3 & 18.1 & 4.0 & 27.5 & 18.4 & 22.5 & 19.9 & 24.1 \\
~ & ViDA & ICLR'24 & 52.9 & 47.9 & 19.4 & 11.4 & 31.3 & 13.3 & 7.6 & 7.6 & 9.9 & 12.5 & 3.8 & 26.3 & 14.4 & 33.9 & 18.2 & 20.7 \\
~ & CMAE & CVPR'24 & 30.6 & 18.9 & 11.5 & 10.4 & 22.5 & 13.9 & 9.8 & 6.6 & 6.5 & 8.8 & 4.0 & 8.5 & 12.7 & 9.2 & 14.4 & 12.6 \\
~ & DPCore & ICML'25 & 22.0 & 18.2 & 14.9 & 14.3 & 24.4 & 13.9 & 12.0 & 11.6 & 10.7 & 15.0 & 5.7 & 21.8 & 15.6 & 12.7 & 18.0 & 15.4 \\
~ & PAID & NeurIPS'25 & 22.9 & 11.8 & 9.9 & 9.1 & 16.7 & 10.8 & 7.4 & 7.4 & 6.6 & 11.4 & 4.5 & 9.3 & 12.8 & 9.4 & 14.5 & 11.0 \\
~ & Ours & Proposed & 15.0 & 10.1 & 7.7 & 6.8 & {12.6} & 5.8 & {4.7} & {4.8} & {4.5} & {4.5} & 2.9 & {3.7} & {8.4} & {5.2} & {9.0} & \textbf{7.0} \\
\bottomrule
\end{tabular}
\end{table*}

\section{Details of Self-Training Loss}\label{L_ST}
Here, we provide the detailed formulation of the self-training loss employed in our online adaptation process. Following the standard teacher-student self-training practice~\cite{Dobler_2023_CVPR, zhu2024reshaping, ni2025maintaining}, we enforce prediction consistency between the teacher and student models on the unlabeled target batch $B_t$ using a symmetric cross-entropy objective:
\begin{equation}
  \mathcal{L}_{ST} = -\sum_{c=1}^{C} q_c\log{p_c} -\sum_{c=1}^{C} p_c\log{q_c},
  \label{eq:self_training}
\end{equation}
where $q$ and $p$ denote the softmax predictions of the teacher and student models, respectively, and $C$ represents the number of classes. Both the teacher model and the student model are initialized with the source model $f_\theta$. Subsequently, the teacher model is continuously updated by the exponential moving average of the student model. 

During adaptation, $\mathcal{L}_{ST}$ is applied exclusively to the unlabeled target data to maintain temporal stability, while the synthetic samples are supervised by our proposed semantic learning objectives. Since the self-training mechanism follows a well-established paradigm and is not the conceptual focus of our contribution, we omit detailed discussion in the main paper and provide its formulation here for reference and reproducibility. 

\paragraph{Compatibility across self-training objectives.}
While we adopt the teacher-student self-training in our main implementation due to its stability, the core contribution of our work, the forward-facilitation via dynamic style bridging, is conceptually orthogonal to the specific choice of the auxiliary self-training objective on the target data. To empirically validate this, we conduct additional experiments replacing the mean-teacher loss with {entropy minimization}~\cite{niu2022efficient}, another widely adopted objective~\cite{niu2022efficient,wang2021tent,han2025ranked} that enforces confidence regularization during adaptation. In this case, our approach updates only the learnable parameters within the normalization layer, rather than the entire model.

As shown in~\cref{tab:self_training_variants}, our framework consistently yields significant improvements over the baseline regardless of the self-training objective employed. Specifically, when integrated with entropy minimization~\cite{niu2022efficient}, our method reduces the average error rate from 49.8\% to 44.8\%. This demonstrates that our proposed synthetic supervision serves as a universal complement to existing self-training techniques, offering robust guidance irrespective of the specific regularization applied to the target data.

\begin{table*}[tb]
\small  
\caption{Semantic segmentation results (mIoU in \%) on the Cityscapes-to-ACDC CTTA task. The four test conditions are repeated three times. All results are evaluated based on the Segformer-B5 architecture. Bold text indicates the best performance.}
\label{tab:segmentation}
\centering
\resizebox{0.93\textwidth}{!}{
\begin{tabular}{l|ccccc|ccccc|ccccc|c} 
\toprule
Time & \multicolumn{15}{c|}{$t\xrightarrow{\makebox[\dimexpr 42\width][c]{\quad}}$} & ~ \\
\midrule
{Round} & \multicolumn{5}{c|}{Round 1} & \multicolumn{5}{c|}{Round 2} & \multicolumn{5}{c|}{Round 3} & All \\ 
\midrule
\multirow{3}*{Condition} & \multirow{3}*{\rotatebox{75}{Fog}} & \multirow{3}*{\rotatebox{75}{Night}} & \multirow{3}*{\rotatebox{75}{Rain}} & \multirow{3}*{\rotatebox{75}{Snow}} & \multirow{3}*{\rotatebox{75}{Mean}} & \multirow{3}*{\rotatebox{75}{Fog}} & \multirow{3}*{\rotatebox{75}{Night}} & \multirow{3}*{\rotatebox{75}{Rain}} & \multirow{3}*{\rotatebox{75}{Snow}} &\multirow{3}*{\rotatebox{75}{Mean}} & \multirow{3}*{\rotatebox{75}{Fog}} & \multirow{3}*{\rotatebox{75}{Night}} & \multirow{3}*{\rotatebox{75}{Rain}} & \multirow{3}*{\rotatebox{75}{Snow}} & \multirow{3}*{\rotatebox{75}{Mean}} & \multirow{3}*{Mean$\uparrow$} \\
~ & ~ & ~ & ~ & ~ & ~ & ~ & ~ & ~ & ~ & ~ & ~ & ~ & ~ & ~ & ~ & ~  \\
~ & ~ & ~ & ~ & ~ & ~ & ~ & ~ & ~ & ~ & ~ & ~ & ~ & ~ & ~ & ~ & ~ \\
\midrule
Source & 69.1 & 40.3 & 59.7 & 57.8 & 56.7 & 69.1 & 40.3 & 59.7 & 57.8 & 56.7 & 69.1 & 40.3 & 59.7 & 57.8 & 56.7 & 56.7 \\
TENT~\cite{wang2021tent} & 69.0 & 40.2 & 60.1 & 57.3 & 56.7 & 68.3 & 39.0 & 60.1 & 56.3 & 55.9 & 67.5 & 37.8 & 59.6 & 55.0 & 55.0 & 55.7 \\
CoTTA~\cite{Wang_2022_CVPR} & 70.9 & 41.2 & 62.4 & 59.7 & 58.6 & 70.9 & 41.1 & 62.6 & 59.7 & 58.6 & 70.9 & 41.0 & 62.7 & 59.7 & 58.6 & 58.6 \\
VDP~\cite{gan2023decorate} & 70.5 & 41.1 & 62.1 & 59.5 & 58.3 & 70.4 & 41.1 & 62.2 & 59.4 & 58.2 & 70.4 & 41.0 & 62.2 & 59.4 & 58.2 & 58.2 \\
SAR~\cite{niu2023towards} & 69.0 & 40.2 & 60.1 & 57.3 & 56.7 & 69.0 & 40.3 & 60.0 & 67.8 & 59.3 & 67.5 & 37.8 & 59.6 & 55.0 & 55.0 & 57.0 \\
ECoTTA~\cite{song2023ecotta} & 68.5 & 35.8 & 62.1 & 57.4 & 56.0 & 68.3 & 35.5 & 62.3 & 57.4 & 55.9 & 68.1 & 35.3 & 62.3 & 57.3 & 55.8 & 55.8 \\
SVDP~\cite{yang2024exploring} & 72.1 & 44.0 & 65.2 & 63.0 & 61.1 & 72.2 & 44.5 & 65.9 & 63.5 & 61.5 & 72.1 & 44.2 & 65.6 & 63.6 & 61.4 & 61.3 \\
OBAO~\cite{zhu2024reshaping} & 71.2 & 42.3 & 65.0 & 62.0 & 60.1 & 72.6 & 43.2 & 66.3 & 63.2 & 61.3 & 72.8 & 43.8 & 66.5 & 63.2 & 61.6 & 61.0 \\

Ours & 71.6 & 43.7 & 66.6 & 64.7 & \textbf{61.7} & 71.2 & 44.6 & 67.1 & 64.4 & \textbf{61.8} & 72.0 & 44.9 & 67.7 & 63.9 & \textbf{62.1} & \textbf{61.9} \\
\bottomrule
\end{tabular}
}
\end{table*}

\begin{table}[th]
\centering
\small
\caption{Statistical reliability analysis. Average online classification error rate (\%) and standard deviation over 5 runs.}
\label{tab:statistical_reliability}
\begin{tabular}{cr}
\toprule
\textbf{Benchmark} & \textbf{Mean Error Rate (\%)} \\
\midrule
ImageNetC & 44.15 $\pm$ 0.16 \\
CIFAR100C & 29.89 $\pm$ 0.13 \\
CIFAR10C & 9.16 $\pm$ 0.09 \\
\bottomrule
\end{tabular}
\end{table}

\section{Additional Experimental Results}\label{additional_results}
\subsection{Statistical Reliability}
\ZLChange{To evaluate the statistical reliability of our method, we repeat the online adaptation process across five independent runs using different random seeds. As shown in~\cref{tab:statistical_reliability}, our method achieves consistent performance across runs, with exceptionally low standard deviations indicating strong stability under stochastic factors. We attribute this robustness to the co-evolving nature of our synthetic knowledge base. This continual evolution ensures a steady stream of reliable, context-aware supervision, effectively suppressing noise accumulation and guaranteeing stable adaptation over time.}

\subsection{Scaling to Different Model Weights} 
To further validate the generalizability and robustness of our proposed framework, we conduct comprehensive evaluations under different pre-trained weight configurations. Following DPCore~\cite{Zhang2025DPCoreDP}, we utilize model weights pre-trained on ImageNet-21k and subsequently fine-tuned on ImageNet-1k. This configuration represents a more sophisticated initialization strategy that leverages broader visual knowledge from the extended ImageNet-21k corpus. We also add comparative methods such as VDP~\cite{gan2023decorate}, ECoTTA~\cite{song2023ecotta}, ViDA~\cite{liu2024vida}, and PAID~\cite{wang2025paid}. For the CIFAR-based experiments, we utilize pre-trained source models provided by prior works~\cite{liu2024continual,liu2024vida,Zhang2025DPCoreDP} to ensure consistency and comparability.

The experimental results are presented in ~\cref{tab:imagenetc_21k_ft_1k,tab:CIFAR_384}, where our method employs identical hyperparameters without additional tuning. Remarkably, in the ImageNet-to-ImageNetC task, our method consistently outperforms all previous approaches, significantly reducing the average classification error rate across 15 domains to 36.1\%, representing a relative improvement of 9.5\% over the current SOTA method DPCore. This substantial improvement demonstrates the effectiveness of our forward-facilitation paradigm in handling diverse and challenging domain shifts. Furthermore, our approach shows particularly strong performance in difficult scenarios such as motion blur and fog noise, where traditional backward-alignment methods often struggle due to the significant stylistic variations these corruptions introduce. These results underscore the practical significance of our framework for real-world deployment scenarios, where models should adapt to unpredictable and diverse environmental conditions while maintaining robust performance.

\begin{table*}[!t]
\caption{Average online classification error rate (\%) over 5 runs in the mixed domains TTA setting.}
\label{tab:inc_mixed}
\small
\setlength{\tabcolsep}{1mm}
\centering
\begin{tabular}{ccccccccc} 
\toprule
Avg. Error (\%) & Source & TENT~\cite{wang2021tent} & EATA~\cite{niu2022efficient} & CoTTA\cite{Wang_2022_CVPR} & SAR~\cite{niu2023towards} & RoTTA~\cite{yuan2023robust} & ROID~\cite{marsden2024universal} & Ours \\
\midrule
ImageNetC & 60.3 & 55.0 & 51.8 & 89.3 & 52.3 & 58.2 & 50.7 & \textbf{44.6}$\pm$0.09 \\
\bottomrule
\end{tabular}
\end{table*}
\begin{table}[!th]
\caption{Average error rate (\%) across various challenging settings on ImageNet-to-ImageNetC.}
\small
\label{more_challenging_settings}
\centering
\resizebox{0.96\linewidth}{!}{
\begin{tabular}{c|c|c|ccccc}
\toprule
\multirow{2}*{Method}  & \multirow{1}*{Class} & \multirow{2}*{CDC} & \multicolumn{5}{c}{Varying Batch Size}  \\
~  &  Imbalance & ~ & 64 & 8 & 4 & 2 & 1 \\
\midrule
DPCore &  43.9 & 42.1 & 39.9 & 43.1 & 45.5 & 78.4 & 82.8 \\
Ours & \textbf{37.6} & \textbf{36.7} & \textbf{36.1} & \textbf{37.1} & \textbf{37.8} & \textbf{40.1} & \textbf{43.4} \\
\bottomrule
\end{tabular}
}
\end{table}

\subsection{Experiments on Segmentation CTTA} 
We extend our evaluation to dense prediction settings using the challenging continual test-time semantic segmentation task Cityscapes-to-ACDC. Following~\cite{zhu2024reshaping,Wang_2022_CVPR}, we employ Segformer-B5~\cite{xie2021segformer} trained on Cityscapes as our segmentation model, and use down-sampled images from ACDC with a resolution of 960×540 as network inputs. Dense prediction tasks require pixel-level semantic annotations, which are not directly obtainable from standard text-to-image models. Therefore, we utilize a small number of synthetic samples from UrbanSyn~\cite{gomez2025all} as the knowledge base. An Adam optimizer with a learning rate of 1e-4 is adopted, and the batch size is set to 1.

The experimental results are summarized in~\cref{tab:segmentation}. We evaluate the performance across three rounds and report the average mIoU metric for each domain and round, providing a comprehensive view of the adaptation effectiveness for repeated domain sequences. The proposed method demonstrates consistent improvement in mIoU across cycles (61.7 $\rightarrow$ 61.8 $\rightarrow$ 62.1), showcasing its capability for long-term adaptation to dynamic environments within dense prediction tasks.

\subsection{Robustness to Mixed Domain Shifts}
\label{sec:mixed_domains}
In real-world scenarios, distribution shifts may not always occur in a temporally continual or gradual manner. To assess the robustness of our framework under varying temporal dynamics, we conduct experiments in a {mixed-domain} setting. Following the protocol in~\cite{Dobler_2023_CVPR,marsden2024universal}, test samples from all 15 ImageNetC corruptions are randomly shuffled to simulate a rapidly changing and unpredictable environment.

Since our style bridging mechanism operates at the instance level rather than relying on batch-level or temporal coherence, it naturally adapts to this setting. As shown in~\cref{tab:inc_mixed}, our method achieves a remarkably low error rate of {44.6\%}, significantly outperforming the state-of-the-art ROID (50.7\%). Notably, methods reliant on temporal continuity (\eg, CoTTA) suffer catastrophic degradation due to negative transfer. In contrast, our performance remains stable compared with the ordered setting, confirming that the style bridging mechanism produces on-the-fly supervision independent of temporal history and ensures robust adaptation regardless of domain sequence.

\subsection{More Challenging Settings}
\ZLChange{We add experiments under \emph{class imbalance}, \emph{CDC shifts}, and \emph{varying batch sizes}. We strictly follow the data stream generation protocol and weight configuration of DPCore~\cite{Zhang2025DPCoreDP} to ensure fair comparison with SOTA. As shown in~\cref{more_challenging_settings}, our method is less batch-hungry and remains stable across all these challenging settings, consistently outperforming the SOTA method in different scenarios.}

\section{Visualization of Synthetic Data}\label{Visualization_Synthetic_Data}
\begin{figure*}[t]
\centering
\includegraphics[width=0.90\textwidth]{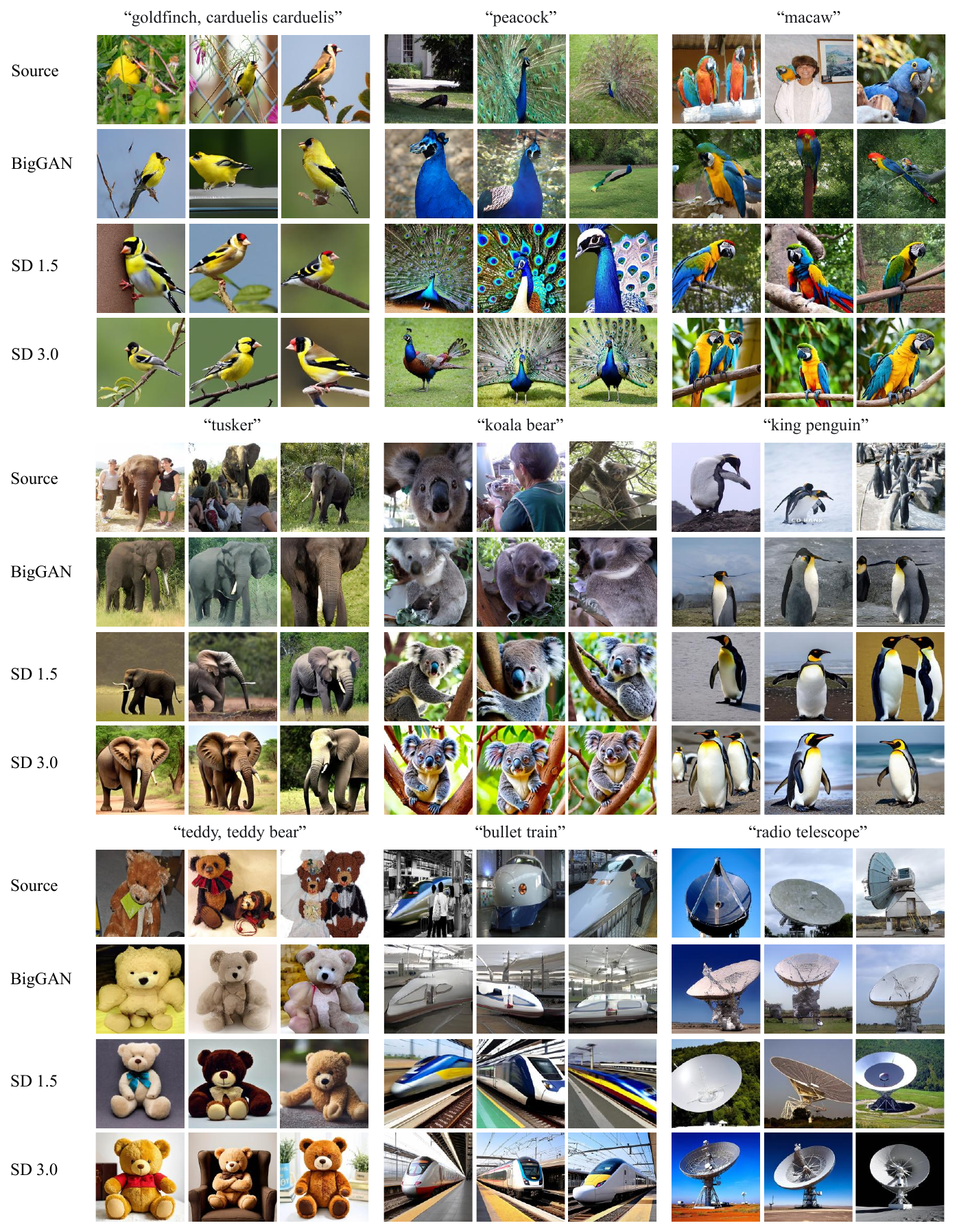} 
\caption{Visualization of synthetic images. We show source images (top row) followed by synthetic samples generated by BigGAN, Stable Diffusion 1.5, and Stable Diffusion 3.0. Each column group contains three randomly selected samples from the same class. Compared to source data, synthetic samples exhibit higher semantic purity (cleaner backgrounds and more salient objects) but also reveal generative bias such as texture overfitting. These observations highlight the necessity of our dynamic style bridging mechanism, which leverages reliable semantics while actively mitigating the inherent generative bias.}
\label{fig:generators_vis}
\end{figure*}

We present visualizations of synthetic images for several ImageNet classes in~\cref{fig:generators_vis}. All images are randomly sampled rather than human-picked and are employed in our experiments. Comparison with real source-domain images reveals consistent patterns across these diverse generators. We observe that while the diffusion-based models generally yield higher fidelity than BigGAN, the synthetic samples across all three generators share a common advantage: they typically exhibit cleaner backgrounds and more prominent foreground subjects compared to the real source domain. This observation strongly validates the motivation behind our ``semantically pure knowledge base", confirming its ability to provide explicit and unambiguous supervision signals. Nonetheless, we also acknowledge the inherent limitations of synthetic data. Compared to the real source domain, the generated samples exhibit a deficiency in diversity, particularly regarding variations in pose and appearance. Moreover, we observe strong model-intrinsic generative biases in certain categories, such as ``koala bear" and ``peacock", where the models tend to overfit to specific textures or canonical compositions.

Crucially, these observations further underscore the necessity of our dynamic style bridging mechanism. They illustrate that naively utilizing these biased samples as static anchors without proper adaptation would inevitably introduce noise and mislead the model to some extent. The consistent superior performance of our framework across multiple generative models demonstrates its ability to \emph{disentangle} reliable semantic content from biased synthetic styles. By effectively bridging the gap between generative bias and target data, our approach circumvents reliance on any particular generative prior, ensuring robust and generalizable adaptation.

\end{document}